\documentclass[12pt]{article}

\usepackage[margin=1.15in]{geometry}
\usepackage[T1]{fontenc}
\usepackage[utf8]{inputenc}
\usepackage{lmodern}
\usepackage{amsmath,amssymb,amsfonts,bm}
\usepackage{booktabs,array,graphicx}
\usepackage[numbers,sort&compress]{natbib}
\usepackage{microtype}
\usepackage{pdflscape}
\usepackage{longtable}
\usepackage{tabularx}
\usepackage{float}
\usepackage[hidelinks]{hyperref}

\title{Ordering Multivariate Time Series by Volatility Memory: A Temporal Convolutional Approach to Volatility Heredity}
\author{}
\date{}

\begin{document}

\begin{center}
{\LARGE Volatility Heredity in Multivariate Time Series: A Statistically Guided Temporal Convolutional Approach to Variable Ordering\\[0.4em]
\par}
\end{center}

\vspace{1.5cm}
\hrule
\vspace{0.8em}

\begin{center}
\begin{minipage}{0.98\linewidth}
\centering

\normalsize
Kaoutar El Azaar\textsuperscript{1,2,3},
Mohamed Hamlich\textsuperscript{1},
Moncef Garouani\textsuperscript{2},\\[0.2em]
Franck Ravat\textsuperscript{2},
Asmae Chakir\textsuperscript{3} and
Mohamed Tabaa\textsuperscript{3}

\par\vspace{0.8em}

\small
\textsuperscript{1} Casablanca (LCCPS Laboratory),
Hassan II University
\par\vspace{0.3em}

\textsuperscript{2} IRIT (UMR 5505 CNRS),
Universit\'e Toulouse Capitole, Toulouse, France
\par\vspace{0.3em}

\textsuperscript{3} LPRI Laboratory, EMSI,
Casablanca, Morocco

\end{minipage}
\end{center}

\begin{abstract}
 Volatility heredity concerns the information that past fluctuations in one series may carry about another after accounting for the target's own persistence. QS-TCN compares this dependence in both directions to estimate an order of temporal precedence. For each series, quasi-maximum likelihood estimation yields a persistence coefficient, and the matrix pencil method extracts temporal modes from the autocovariances of squared observations. These estimates define filters in a temporal convolutional network. The network models the target's squared observations from their own past and scores the association between the residuals and the filtered history of a candidate source. Differences between the two directional scores are aggregated into a global order.

In an exploratory simulation with 100 systems in each of four volatility regimes, an ensemble of ten models trained with fixed seeds achieved higher mean ordering quality than CCF-AUC, a lead--lag method based on distance correlation, in every regime. Against CCF-AUC applied to squared standardized observations, the mean gains were 8.69 percentage points under positive linear coupling and 7.81 points when nonlinear coupling depended on the target's state. In the other two regimes, paired 95\% bootstrap intervals for the gains included zero. Mean agreement with reference constraints was 71.55\% across nine Finance systems and 54.02\% across 27 simulated functional magnetic resonance imaging systems, compared with 50\% expected under random ordering. In a panel of daily currency and gold returns, the estimated order placed the dollar--franc and dollar--yen exchange rates before gold.
\end{abstract}


\vspace{0.8em}
\hrule

\section{Introduction}
\label{sec:introduction}

Understanding the directional relationships between the variables of a multivariate time series is important in finance and neuroscience. In financial markets, the fluctuations of one asset can precede those of another. In neuroscience, temporal relationships between recorded signals help characterize interactions between brain regions. In both cases, a useful goal is to organize the variables in a global order that summarizes the directional evidence present in the data.

Several approaches address related questions from different perspectives. Granger~\citep{granger1969} defines direction through the additional predictive information provided by a variable's past. Graphical methods rely on conditional independence relations to study dependencies among several variables, including nonlinear relationships in time series~\citep{spirtes2000,eichler2012,runge2019}. Other approaches estimate structural equation models through continuous optimization~\citep{pamfil2020} or exploit assumptions on the noise distribution to identify causal structure. LiNGAM, for example, relies on non-Gaussian disturbances~\citep{shimizu2006}. The method has been extended to time series~\citep{hyvarinen2010}, with further developments addressing functional MRI~\citep{xu2014} and functional data~\citep{yang2022}. These approaches offer different ways of extracting directional information from observational data.

Volatility provides another source of temporal information. Econometric research has developed causality-in-variance tests~\citep{cheung1996,hong2001,hafner2006,hafner2008}, models of interdependent conditional variances~\citep{engle1995}, and measures of volatility transmission between markets~\citep{diebold2012}. In causal discovery, location-scale noise models provide conditions under which causal direction is identifiable from heteroscedastic observational data~\citep{immer2023}. More recent work distinguishes causal effects on the mean from those on the variance by learning separate causal graphs from heteroscedastic observational data~\citep{chikahara2026}. Building on this literature, the present study examines how estimates of volatility memory can contribute to learning an order among variables.

The persistence of fluctuations alone does not establish a direction. A variable can display several temporal scales through its own dynamics, while its dependencies on other variables can add further structure to the observed series. Results on Prony systems and related super-resolution problems show how closely spaced components can make parameter recovery sensitive to measurement errors~\citep{batenkov2013,batenkov2021}. These difficulties motivate an approach that first represents the target variable's history, then examines the relationship between the source's past and the remaining fluctuations. The target's history provides a predictive representation, without being interpreted as an isolated structural mechanism.

This article proposes QS-TCN, a two-channel method that connects statistical estimates of volatility memory to learning with temporal convolutions. The first channel estimates persistence by quasi-maximum likelihood under a working stochastic-volatility model~\citep{harvey1994multivariate,ruiz1994quasi} and extracts temporal modes using the Matrix Pencil method~\citep{hua1990matrix,sarkar1995using}. In the second channel, the persistence estimates and accepted poles define structured temporal filters, which are combined with learned filters. The network represents the target's past and evaluates directional evidence associated with its residual dynamics. A shared scoring function evaluates the two orientations of each pair, and the resulting contrast is exactly antisymmetric. Synchronized block-bootstrap samples allow the stability of the scores to be assessed, and a decoder combines them into a global order. This order summarizes directional evidence and does not identify direct causal links.

The evaluation covers controlled synthetic systems with different volatility mechanisms and the Finance and fMRI datasets, for which reference directional constraints are available~\citep{nauta2019,kleinberg2012causality,smith2011}. In the synthetic experiments, QS-TCN is compared with an ordering constructed from the lead--lag scores of \citet{bennett2022leadlag}. On the Finance and fMRI datasets, it is compared with the contemporaneous order returned by VAR-LiNGAM~\citep{hyvarinen2010}. These comparisons concern agreement between the recovered orders and the reference directional constraints. The methods are not treated as equivalent procedures for graph reconstruction. Applications to currency, gold and cryptocurrency markets extend the study to observed financial data.

The evaluation reports ordering quality for each system and for the datasets as a whole, together with its sensitivity to resampling and the agreement expected from a random order. Simple marginal properties can also align with reference directional constraints~\citep{reisach2021,lohse2025}. The evaluation of the full model is therefore distinguished from the attribution of its performance to individual components.

This work makes three contributions. The first is a framework for ordering the variables of multivariate time series using temporal information in volatility dynamics. The second connects QML persistence estimates and Matrix Pencil modes to learned temporal filters, with a shared directional scoring function and a global ordering procedure. The third is an evaluation across controlled volatility mechanisms, the Finance and fMRI datasets, and observed financial markets, including external comparisons and a bootstrap sensitivity analysis.

Section~\ref{sec:background} presents the main technical difficulties and evaluation issues. Section~\ref{sec:methodology} describes QS-TCN, Section~\ref{sec:results} reports the experiments and results, and Section~\ref{sec:conclusion} concludes.

\section{Background}
\label{sec:background}

Volatility-based ordering examines temporal information in the amplitude of fluctuations. Interpreting this information requires distinguishing conditional variance from the conditional second moment. For a variable $X_t$ with finite second moment and an available history $\mathcal H_{t-1}$,
\begin{equation}
\mathbb E[X_t^2\mid\mathcal H_{t-1}]
=
\operatorname{Var}(X_t\mid\mathcal H_{t-1})
+
\bigl(\mathbb E[X_t\mid\mathcal H_{t-1}]\bigr)^2.
\label{eq:conditional-second-moment}
\end{equation}
Dependencies between squared observations can therefore reflect both mean and variance dynamics. This distinction matters for directional analysis, since neglecting dependence in the conditional mean can distort tests for causality in variance~\citep{pantelidis2004}.

A second difficulty concerns the temporal structure generated by volatility persistence. In the stationary stochastic-volatility reference model of Section~\ref{sec:channel1}, Gaussian log-volatility follows a first-order autoregression with a single persistence parameter $\phi$. Nevertheless, the positive-lag autocovariance of the squared observations admits an expansion in geometric components with rates $\phi,\phi^2,\phi^3,\ldots$, as derived in Appendix~\ref{app:sv-modes}. Multiple temporal modes can thus arise within a univariate model, without cross-variable coupling. An additional estimated mode does not, by itself, identify an external influence.

Over a finite lag range, this memory profile can be approximated by a short sum of exponentials whose parameters are estimated using the Matrix Pencil method~\citep{hua1990matrix,sarkar1995using}. Separating nearby components remains sensitive to estimation error, a difficulty studied in Prony systems and related super-resolution problems~\citep{batenkov2013,batenkov2021}. Persistence estimates and accepted poles therefore provide summaries of marginal temporal structure. In QS-TCN, these summaries define structured filters that connect Channel~I to the temporal representations learned in Channel~II, without attributing individual poles to specific causal mechanisms.

Pairwise directional scores do not necessarily define a consistent global order. Antisymmetry ensures that reversing a pair reverses its score, but does not ensure transitivity. For example, $S_{ij}>0$, $S_{jk}>0$ and $S_{ki}>0$ form a directed cycle. Ranking methods must reconcile such inconsistencies when aggregating pairwise comparisons~\citep{jiang2011}. Global decoding therefore seeks an order that reflects the combined evidence, which can require placing some pairs against their individually preferred orientation.

Evaluation must also distinguish agreement with a reference structure from the information used to obtain that agreement. Marginal properties, including variance, can align with reference structures and account for strong performance on some synthetic benchmarks~\citep{reisach2021}. Related sortability effects occur in time-series settings~\citep{lohse2025}. These findings motivate comparisons that assess whether pairwise temporal representations contribute information beyond node-specific summaries.

The fMRI setting introduces an additional observation layer. Neural activity is measured indirectly through the haemodynamic response, with a temporal resolution that can be coarser than the neural interactions of interest. Results for Granger-type analysis identify conditions under which haemodynamic filtering preserves directional relationships, while also examining the effects of downsampling and measurement noise~\citep{seth2013,novelli2025}. The NetSim benchmark provides simulated BOLD data for evaluating network methods under different observation conditions~\citep{smith2011}. It therefore offers a controlled setting for examining how informative volatility-based temporal representations remain after haemodynamic transformation and limited temporal sampling.

\section{Methodology}
\label{sec:methodology}

Given a multivariate time series $\mathbf X_t\in\mathbb R^d$, QS-TCN estimates a permutation $\widehat\pi=(\widehat\pi_1,\ldots,\widehat\pi_d)$, in which $\widehat\pi_a$ is the variable placed in position $a$, describing directional precedence among its variables. It combines two channels: a statistical description of marginal volatility memory and a temporal network that converts this information into pairwise scores. These scores are subsequently aggregated into a global order. The output specifies precedence, not the existence of a direct causal edge.

\subsection{Channel I: QML persistence and spectral memory}
\label{sec:channel1}

For each variable, we use the stationary stochastic-volatility reference
\begin{equation}
 X_{i,t}=\exp(h_{i,t}/2)\varepsilon_{i,t},
 \qquad
 h_{i,t}=\mu_i+\phi_i(h_{i,t-1}-\mu_i)+\eta_{i,t},
 \label{eq:sv-reference}
\end{equation}
where $|\phi_i|<1$ and the innovation sequences are independent and Gaussian. Following \citep{harvey1994multivariate,ruiz1994quasi}, the persistence $\phi_i$ is estimated by QML: squared observations are log-transformed, and a Kalman filter evaluates a Gaussian approximation to their likelihood. The estimate characterizes marginal persistence rather than an identified structural noise process.

To represent additional temporal scales, we form the centered-square signal $q_{i,t}=X_{i,t}^2-\overline{X_i^2}$ and approximate its positive-lag autocovariance by
\begin{equation}
 \widehat{\mathcal C}_i(k)
 \approx\sum_{r=1}^{R_i}A_{ir}\lambda_{ir}^{k-1},
 \qquad k=1,\ldots,K_{\mathrm{spec}}.
 \label{eq:profile}
\end{equation}
The Matrix Pencil method of \citep{sarkar1995using} extracts the poles from a reduced singular subspace of the profile's Hankel matrix. A block-bootstrap threshold determines the supported rank, and amplitudes are fitted by least squares. Channel~II receives the numerical persistence $\widetilde\phi_i$, the accepted real poles, and a descriptor $\Theta_i$ summarizing estimation diagnostics. Unsupported modes are masked. The decomposition concerns the total squared-memory profile; individual poles are not assigned a causal origin.

\subsection{Channel II: memory--heredity temporal network}
\label{sec:channel2}

Channel~II uses shared causal temporal convolutions to represent lagged dependencies, drawing on their application to multivariate time-series forecasting \citep{borovykh2019dilated}. The persistence estimates $\widetilde\phi_i$ and accepted poles $\widehat\Lambda_i$ from Channel~I enter the TCN through structured convolution kernels, which are combined with learned temporal filters. These estimated rates remain fixed during neural training.

Let $\mathbf q_{j,t}^{(L)}=(q_{j,t-1},\ldots,q_{j,t-L})^\top$ denote a lagged input window.

The \emph{Memory Cell} represents the target's own history through free filters and geometric filters proportional to $\widetilde\phi_j^{\ell-1}$ and $\widehat\lambda_{jr}^{\ell-1}$, using the target's persistence and accepted poles. A nonlinear readout produces
\begin{equation}
 \widehat q_{j,t}
 =\operatorname{Mem}_{\theta_m}
 \bigl(\mathbf q_{j,t}^{(L_m)};\widetilde\phi_j,\widehat\Lambda_j,\Theta_j\bigr),
 \qquad r_{j,t}=q_{j,t}-\widehat q_{j,t}.
 \label{eq:memory}
\end{equation}
The residual captures fluctuations left unexplained by this fitted memory representation.

For the direction $i\to j$, the \emph{Heredity Cell} filters the source history using the source persistence $\widetilde\phi_i$ and the target poles $\widehat\lambda_{jr}$. Its structured kernels are
\begin{equation}
 K_{ijr}(\ell)=
 \sum_{s=0}^{\ell-1}\widetilde\phi_i^{\ell-1-s}
 \widehat\lambda_{jr}^{s},
 \qquad \ell=1,\ldots,L_h.
 \label{eq:heredity-kernel}
\end{equation}
These kernels are normalized and combined with free source filters through learned weights. The representation retains the temporal means of both branches and their mean products with $r_{j,t}$, together with $\Theta_i$ and $\Theta_j$. The same scoring network evaluates this representation in both orientations:
\begin{equation}
 D_{ij}=f_\theta(i,j)-f_\theta(j,i),
 \qquad
 S_{ij}=S_{0,ij}+g_{ij}D_{ij},
 \quad g_{ij}=g_{ji}.
 \label{eq:score}
\end{equation}
Here $S_{0,ij}$ is an analytical anchor comparing lagged covariances between the source's squared signal and autoregressive residuals of the target. A symmetric learned gate $g_{ij}$ controls the correction, giving $S_{ji}=-S_{ij}$ exactly. The kernel itself is symmetric in its rates; directional evidence arises from the source--target assignment and the residual-based comparison.

\subsection{Learning and global ordering}
\label{sec:training}

The network is trained on synthetic systems with acyclic coupling in latent log-volatility. The Memory Cell first minimizes squared prediction error and is then frozen. The directional module minimizes a logistic precedence loss, with a penalty on deviations from the anchor. Its checkpoint is selected on independent synthetic validation systems; benchmark graph labels are excluded from optimization and checkpoint selection.

Benchmark inference uses a fixed chronological prefix, also used for centering, scaling, and Channel-I estimation. The market applications use the full series or windows specified below. Neural parameters remain fixed. Synchronized block-bootstrap views preserve alignment across variables, and the anchor is refitted within each view. Pairwise scores are aggregated by their median, weighted by sign stability and penalized for dispersion, yielding an antisymmetric matrix $C$. These weights describe resampling stability rather than causal probabilities.

The decoder seeks an order with high agreement with this matrix:
\begin{equation}
 J(\pi;C)=\sum_{1\le a<b\le d}C_{\pi_a\pi_b}.
 \label{eq:order-objective}
\end{equation}
Starting from decreasing row sums, it accepts improving single-node insertions until no improvement is found or the limit of 25 moves is reached. The returned order is therefore a heuristic estimate. Appendix~\ref{app:implementation} specifies the structured filters, statistical score $S_0$, bootstrap aggregation and training settings used in the controlled synthetic evaluation.

\section{Results}
\label{sec:results}

This section asks how well QS-TCN orders variables when it is applied exactly as frozen.
Two questions guide the analysis.
How closely does the returned order agree with the structure known for each dataset, and how stable is the resulting OQ under resampling?
The analysis starts with two simulated benchmarks, Finance and fMRI, whose reference structure allows a direct measurement of agreement.
A controlled synthetic evaluation follows (Section~\ref{sec:external-leadlag}).
The last part turns to real markets, where no complete reference order exists.
It covers a panel of six currency pairs and gold, then several cryptocurrency systems whose recovered orders are compared with relations published by other authors.

The Finance, fMRI, Forex/Gold and cryptocurrency experiments all use the frozen implementation without modification.
The controlled synthetic evaluation differs, since it uses frozen checkpoints of a reconstructed implementation (Section~\ref{sec:external-leadlag}).
On Finance and fMRI, inference is repeated under ten bootstrap seed schemes labelled A to J, and scheme A is the historical frozen run.
All ten recorded schemes are reported, and the neural parameters stay fixed during evaluation.
One point needs attention from the start.
The variation across seed schemes measures the sensitivity of the frozen inference to the bootstrap construction.
This variation is neither an estimate of training uncertainty nor an estimate of sampling uncertainty.

\subsection{Datasets and evaluation setting}

Table~\ref{tab:datasets} lists the four benchmark and market settings.
The controlled synthetic systems are described separately in Section~\ref{sec:external-leadlag}.

The Finance data are the preprocessed financial benchmark distributed with TCDF \citep{nauta2019}, which comes from the simulated financial systems of Kleinberg \citep{kleinberg2012causality}.
The fMRI data are the preprocessed NetSim benchmark, also distributed with TCDF \citep{nauta2019} and first introduced by Smith et al.~\citep{smith2011}.
In NetSim, the directed neural activity is not observed directly.
This activity passes through a haemodynamic transformation, so the analysed BOLD signals combine temporal smoothing, nonlinear observation effects and coloured noise \citep{friston2000}.

The original NetSim benchmark contains 28 systems.
Dataset~21 is missing from the retained frozen evaluation output, and the retained files do not explain why.
Every fMRI aggregate below therefore covers the 27 available systems.
A later byte-level check showed that the observations and the reference-edge file of dataset~21 are identical to those of dataset~1.
The original reason for the omission was not recorded.

The two other settings use real financial data.
The Forex/Gold panel has seven assets and shows what QS-TCN returns when no complete reference graph exists.
The cryptocurrency experiments serve a different purpose.
Their recovered orders are compared with directional structures published by other authors: the six-cryptocurrency connectedness analysis of Ji et al.~\citep{ji2019} and the regime-specific spillover analysis of Bitcoin, Ethereum and Litecoin by Yousaf and Ali~\citep{yousafali2020}.
These published structures serve as points of comparison and are not treated as causal ground truth.

\begin{table*}[t]
\centering
\caption{Datasets used in the empirical evaluation. A known reference structure allows direct computation of ordering quality. Published market relations are used only for cross-study comparison and are not interpreted as complete causal ground truth.}
\label{tab:datasets}
\scriptsize
\setlength{\tabcolsep}{3.5pt}
\renewcommand{\arraystretch}{0.94}
\resizebox{0.94\textwidth}{!}{%
\begin{tabular}{lllll}
\toprule
Data setting & Domain & Systems / assets & Length & Reference structure   \\
\midrule
Finance~\citep{nauta2019,kleinberg2012causality} & Simulated financial systems & 9 datasets, 25 variables & $T=4000$ & Known   \\
fMRI~\citep{nauta2019,smith2011} & Simulated BOLD systems & 27 evaluated datasets, 5--50 nodes & $T=50$--$5000$ & Known   \\
Forex/Gold & Observed financial markets & 7 assets & $T=2384$ & No complete GT  \\
Cryptocurrencies~\citep{ji2019,yousafali2020} & Observed crypto markets & 3 or 6 assets & Study-dependent & Published directional relations   \\
\bottomrule
\end{tabular}}
\end{table*}

The benchmark analyses have a clear scope.
They measure how well the recovered order agrees with the reference constraints and how stable the OQ is under the bootstrap.
They do not show that the recovered signal comes only from volatility.
A descriptive VAR-LiNGAM comparison is therefore included below, whereas marginal-ordering baselines and ablations on the benchmark data are absent.
The synthetic evaluation compares QS-TCN with an external lead-lag ordering method (Section~\ref{sec:external-leadlag}).

\paragraph{Ordering quality.}
A benchmark states, for some pairs of variables, which one must come first.
The score is the share of these requirements which the returned order respects.
The precise definition follows.
Fix a dataset $m$ and call its variables nodes.
Let $\mathcal A_m$ be the set of directed reference constraints of dataset $m$.
A pair $(i,j)\in\mathcal A_m$, with $i\neq j$, means that the reference structure requires node $i$ to precede node $j$ for the purpose of the score.
This definition does not require a single reference permutation.
The recovered order is a permutation $\widehat\pi_m$ of the nodes, and $p_{\widehat\pi_m}(i)$ denotes the position of node $i$ in this permutation.
For a nonempty reference set, the ordering quality of dataset $m$ is
\begin{equation}
\mathrm{OQ}_m
=
\frac{1}{|\mathcal A_m|}
\sum_{(i,j)\in\mathcal A_m}
\mathbb I
\left[
p_{\widehat\pi_m}(i)
<
p_{\widehat\pi_m}(j)
\right].
\label{eq:oq-results}
\end{equation}
In this formula, $|\mathcal A_m|$ is the number of scored constraints, and the indicator $\mathbb I[\cdot]$ equals $1$ when the statement inside the brackets is true and $0$ otherwise.
Thus $\mathrm{OQ}_m$ lies between $0$ and $1$.
The value $1$ means that every constraint is respected, and the value $0$ means that none is.

Each benchmark supplies $\mathcal A_m$ as the list of directed edges of its ground-truth graph.
Self-loops are removed, since a node does not precede itself, and no transitive closure is applied, so only direct edges are scored.
Under this construction, the benchmarks contain $197$ constraints over the nine Finance datasets and $239$ over the $27$ evaluated fMRI datasets, and these are the totals used below.
Some reference graphs contain a bidirectional pair, meaning that both $(i,j)$ and $(j,i)$ belong to $\mathcal A_m$, or a longer directed cycle.
No linear order respects both constraints of such a pair, or all constraints of a directed cycle, so $\mathrm{OQ}_m$ stays below $1$ for every method.
Enumerating the orders inside each strongly connected component and arranging the components in a topological order gives the exact maximum attainable value.
Rounded to three decimal places, this maximum is $0.625$ for fMRI dataset~13, $0.800$ for fMRI dataset~14, $0.947$ for Finance 20\_1D, $0.950$ for 20\_1\_3 and for 40\_1\_3, and $0.974$ for 40\_1.
All other datasets have an acyclic reference graph, and their maximum is $1$.
The value $0.625$ reported for dataset~13 under every seed scheme therefore equals its maximum.

A small example, which is not data, makes the computation concrete.
Take three nodes and the constraints $(1,2)$ and $(2,3)$.
The order $(2,1,3)$ respects $(2,3)$ and violates $(1,2)$, so $\mathrm{OQ}=1/2$.

\paragraph{What a random order obtains.}
A random order gives the reference point for these scores.
Suppose the order is drawn uniformly among all permutations of the nodes.
For a fixed pair $i\neq j$, exchanging the roles of $i$ and $j$ maps the permutations with $p(i)<p(j)$ one-to-one onto those with $p(j)<p(i)$, so each event has probability $1/2$.
By linearity of expectation, $\mathbb E[\mathrm{OQ}_m]=1/2$ for every nonempty set $\mathcal A_m$, whatever the dependence between constraints.
Dependence matters for the distribution of $\mathrm{OQ}_m$ and not for its mean.
In the example above, the six permutations give the values $1$, $1/2$, $1/2$, $1/2$, $1/2$ and $0$.
Both constraints are respected with probability $1/6$, whereas independence would give $1/4$.
Dependence also works in the other direction.
If node $i$ must precede both $j$ and $k$, both constraints are respected exactly when $i$ comes first among the three nodes, which has probability $1/3$.
For this reason, a binomial calculation which treats all scored pairs as independent is not a valid significance test, and none is used here.

\paragraph{Two ways of summarizing a benchmark.}
Finance and fMRI are evaluated under $S=10$ seed schemes.
Write $\mathrm{OQ}_m^{(s)}$ for the value of dataset $m$ under scheme $s$, and $\overline{\mathrm{OQ}}_m=S^{-1}\sum_{s}\mathrm{OQ}_m^{(s)}$ for its average over schemes, which is the ``Mean OQ'' column of the tables.
With $M$ datasets, the macro-average gives the same weight to every dataset,
\begin{equation}
\mathrm{OQ}_{\mathrm{macro}}
=
\frac{1}{M}
\sum_{m=1}^{M}\overline{\mathrm{OQ}}_m,
\end{equation}
and the pair-weighted score gives the same weight to every scored constraint, so datasets with more constraints count more,
\begin{equation}
\mathrm{OQ}_{\mathrm{weighted}}
=
\frac{
\sum_{m=1}^{M}
|\mathcal A_m|\,\overline{\mathrm{OQ}}_m
}{
\sum_{m=1}^{M}
|\mathcal A_m|
}.
\end{equation}
Two standard deviations appear below, and they answer different questions.
The first is computed across schemes.
If $\mathrm{OQ}^{(s)}_{\mathrm{macro}}$ is the macro-average of scheme $s$, then
\begin{equation}
\mathrm{SD}_{\mathrm{seed}}
=
\left[
\frac{1}{S-1}
\sum_{s=1}^{S}
\left(\mathrm{OQ}^{(s)}_{\mathrm{macro}}-\mathrm{OQ}_{\mathrm{macro}}\right)^2
\right]^{1/2}.
\end{equation}
This quantity describes how much the benchmark average moves when the bootstrap seed scheme changes.
The second is computed across datasets,
\begin{equation}
\mathrm{SD}_{\mathrm{datasets}}
=
\left[
\frac{1}{M-1}
\sum_{m=1}^{M}
\left(\overline{\mathrm{OQ}}_m-\mathrm{OQ}_{\mathrm{macro}}\right)^2
\right]^{1/2},
\end{equation}
and describes how different the datasets are from one another.
Every standard deviation in this section uses the denominator $n-1$.
Neither quantity is a confidence interval for performance on new systems.

For the real markets, no benchmark OQ is reported, since no complete reference structure exists.
The recovered order is reported instead.
When a published directional structure maps onto the same assets, concordance is measured separately from benchmark ordering quality.

\paragraph{VAR-LiNGAM comparison.}
VAR-LiNGAM serves as a competing temporal-ordering method.
The upstream implementation is fitted to the same chronological estimation prefix used by QS-TCN, with $T_0=\min\{T,\max\{40,\lfloor0.65T\rfloor\}\}$, one VAR lag, and pruning enabled~\citep{hyvarinen2010}.
With the maximum lag fixed at one, the BIC setting has no alternative lag order to select.
No reference graph is supplied during fitting.
The resulting contemporaneous order is scored against the same direct reference constraints used for QS-TCN, after omitting their delay labels.
The comparison is therefore descriptive and is not a like-for-like comparison of lag-specific graph recovery.
Tables~\ref{tab:finance-detailed} and~\ref{tab:fmri-detailed} give the per-system values, and the full seed-wise QS-TCN scores appear in Appendix~\ref{app:seed-results}.

\subsection{Finance benchmark}

The Finance benchmark distributed with TCDF \citep{nauta2019} contains nine evaluated systems derived from the financial simulations of Kleinberg \citep{kleinberg2012causality}.
Averaging the nine dataset-level values over the ten seed schemes gives
\begin{equation}
    \mathrm{OQ}_{\mathrm{Finance,macro}}=0.7155.
\end{equation}
Across the ten schemes, the macro-average has a standard deviation $\mathrm{SD}_{\mathrm{seed}}=0.0156$, and the scheme-level macro-averages range from
\begin{equation}
    0.6961\le\mathrm{OQ}^{(s)}_{\mathrm{Finance,macro}}\le0.7465.
\end{equation}
The value $0.0156$ describes only the sensitivity of the benchmark mean to the seed scheme.
It is not the variation between Finance systems, which is much larger.
From the dataset means of Table~\ref{tab:finance-detailed}, the calculation gives $\mathrm{SD}_{\mathrm{datasets}}\approx0.138$.
Giving the same weight to every scored reference pair instead of every dataset yields
\begin{equation}
    \mathrm{OQ}_{\mathrm{Finance,weighted}}
    =
    0.6914
\end{equation}
over $197$ scored reference pairs.
The historical frozen scheme gives a macro-OQ of $0.7193$, close to the ten-scheme macro-average.
On the same nine systems and prefixes, VAR-LiNGAM obtains macro and pair-weighted OQ values of $0.5989$ and $0.5584$.
Its macro score is below the historical QS-TCN score.
The comparison remains descriptive, because VAR-LiNGAM supplies one deterministic contemporaneous order whereas the scoring constraints include lagged relations.

\begin{table*}[t]
\centering
\caption{Finance ordering quality by dataset. QS-TCN A is the historical frozen order; VAR-LiNGAM is one deterministic order fitted on the same estimation prefix. Mean, standard deviation and range for QS-TCN are computed across the ten bootstrap seed schemes A--J.}
\label{tab:finance-detailed}
\scriptsize
\setlength{\tabcolsep}{4pt}
\renewcommand{\arraystretch}{0.94}
\resizebox{0.96\textwidth}{!}{%
\begin{tabular}{l r c c c c c}
\toprule
Dataset & Pairs & QS-TCN A & VAR-LiNGAM & Mean OQ & SD & Min--Max \\
\midrule
manyinputs & 6 & 1.000 & 1.000 & 1.000 & 0.000 & 1.000--1.000 \\
20\_1A & 18 & 0.778 & 0.500 & 0.806 & 0.054 & 0.722--0.889 \\
20\_1B & 20 & 0.650 & 0.650 & 0.590 & 0.052 & 0.500--0.650 \\
20\_1C & 19 & 0.526 & 0.421 & 0.532 & 0.017 & 0.526--0.579 \\
20\_1D & 19 & 0.684 & 0.474 & 0.705 & 0.037 & 0.632--0.737 \\
20\_1E & 17 & 0.824 & 0.765 & 0.782 & 0.088 & 0.647--0.882 \\
40\_1 & 38 & 0.737 & 0.605 & 0.700 & 0.051 & 0.605--0.763 \\
40\_1\_3 & 40 & 0.725 & 0.475 & 0.700 & 0.051 & 0.625--0.775 \\
20\_1\_3 & 20 & 0.550 & 0.500 & 0.625 & 0.054 & 0.550--0.750 \\
\midrule
Mean over datasets & -- & \textbf{0.7193} & 0.5989 & 0.7155 & \multicolumn{2}{c}{$\mathrm{SD}_{\rm seed}=0.0156$} \\
\bottomrule
\end{tabular}}
\end{table*}

The highest dataset-level mean is on \texttt{manyinputs}, where all six scored reference constraints are oriented correctly under every seed scheme,
\begin{equation}
    \mathrm{OQ}_{\mathrm{manyinputs}}=1.
\end{equation}
Only six directed constraints are scored for this system.
The value therefore means perfect agreement on these six admissible pairs and does not mean that every possible pairwise relation among the 25 variables is recovered.
The systems \texttt{20\_1A} ($0.806$) and \texttt{20\_1E} ($0.782$) also show high mean OQ.
These values are descriptive and are not presented as dataset-wise significance claims, since no graph-specific random-order distribution has been evaluated.

The weaker systems show a different pattern.
Take \texttt{20\_1C}.
Its mean OQ is $0.532$ and its seed-scheme SD is $0.017$, the smallest nonzero value among the Finance systems.
Its agreement with the reference constraints is therefore stable across the ten seed schemes, at a level close to the random-order expectation of $0.5$.
A larger seed-scheme SD would instead show that OQ is sensitive to the bootstrap construction.
These two properties are separate.
Seed-scheme stability describes the reproducibility of the ordering score, and OQ measures agreement with the reference.
Equal OQ values do not imply identical recovered orders.
On Finance, QS-TCN reaches a macro-OQ of $0.7155$ and a pair-weighted OQ of $0.6914$, both above the random-order expectation of $0.5$, together with large differences between individual systems.
The per-system VAR-LiNGAM comparison appears in Table~\ref{tab:finance-detailed}, and no marginal-statistic baseline is included.

\subsection{fMRI benchmark}

The fMRI experiments use the NetSim systems of Smith et al.~\citep{smith2011}, distributed in preprocessed form with TCDF \citep{nauta2019}.
These simulations generate directed neural activity and then observe it through BOLD dynamics, which adds temporal smoothing and nonlinear haemodynamic effects \citep{friston2000}.
Twenty-seven datasets are present in the frozen QS-TCN evaluation, and all fMRI aggregates below are restricted to these systems, as stated in the dataset description above.

Averaging the 27 dataset-level values over the ten seed schemes gives
\begin{equation}
    \mathrm{OQ}_{\mathrm{fMRI,macro}}=0.5402.
\end{equation}
Its seed-scheme standard deviation is $\mathrm{SD}_{\mathrm{seed}}=0.0187$, and the scheme-level values range from
\begin{equation}
    0.5142\le\mathrm{OQ}^{(s)}_{\mathrm{fMRI,macro}}\le0.5626.
\end{equation}
As for Finance, $0.0187$ describes the sensitivity of the benchmark mean to the seed scheme and is not a confidence interval.
From the dataset means of Table~\ref{tab:fmri-detailed}, the calculation gives $\mathrm{SD}_{\mathrm{datasets}}\approx0.141$.
Giving the same weight to each of the $239$ scored reference constraints yields
\begin{equation}
    \mathrm{OQ}_{\mathrm{fMRI,weighted}}
    =
    0.5251.
\end{equation}
Equivalently, 1255 of the $2390$ constraint evaluations ($239$ constraints in each of the ten schemes) are correctly oriented.
A uniformly random permutation has an expected OQ of $0.5$, so the pair-weighted fMRI result lies only slightly above this expectation.
Without a graph-level permutation analysis, the aggregate fMRI score is not interpreted as a statistically established departure from random ordering.
The historical frozen scheme gives a macro-OQ of $0.5236$.
On the same 27 systems and estimation prefixes, VAR-LiNGAM obtains macro and pair-weighted OQ values of $0.3764$ and $0.4100$.
These scores are lower than the historical QS-TCN aggregate.
The contemporaneous-versus-lagged difference limits the interpretation, and the fMRI benchmark remains highly heterogeneous across datasets.

\begin{table*}[t]
\centering
\caption{fMRI ordering quality by dataset. QS-TCN A is the historical frozen order; VAR-LiNGAM is one deterministic order fitted on the same estimation prefix. Mean, standard deviation and range for QS-TCN are computed across the ten bootstrap seed schemes A--J.}
\label{tab:fmri-detailed}
\scriptsize
\setlength{\tabcolsep}{3.6pt}
\renewcommand{\arraystretch}{0.92}
\resizebox{0.96\textwidth}{!}{%
\begin{tabular}{c r c c c c c}
\toprule
Dataset & Pairs & QS-TCN A & VAR-LiNGAM & Mean OQ & SD & Min--Max \\
\midrule
1&5&0.400 & 0.400&0.460&0.165&0.200--0.600\\
2&11&0.545 & 0.727&0.618&0.120&0.455--0.818\\
3&18&0.444 & 0.444&0.439&0.032&0.389--0.500\\
4&61&0.475 & 0.475&0.503&0.021&0.475--0.525\\
5&5&0.400 & 0.200&0.380&0.063&0.200--0.400\\
6&11&0.273 & 0.273&0.318&0.048&0.273--0.364\\
7&5&0.400 & 0.000&0.400&0.000&0.400--0.400\\
8&5&0.800 & 0.800&0.780&0.063&0.600--0.800\\
9&5&0.800 & 0.400&0.780&0.063&0.600--0.800\\
10&5&0.600 & 0.400&0.600&0.163&0.200--0.800\\
11&11&0.545 & 0.455&0.636&0.096&0.455--0.727\\
12&11&0.727 & 0.455&0.645&0.125&0.455--0.818\\
13&8&0.625 & 0.375&0.625&0.000&0.625--0.625\\
14&5&0.400 & 0.200&0.640&0.158&0.400--0.800\\
15&5&0.600 & 0.000&0.560&0.084&0.400--0.600\\
16&7&0.429 & 0.286&0.443&0.171&0.143--0.714\\
17&11&0.273 & 0.273&0.436&0.103&0.273--0.545\\
18&5&0.600 & 0.600&0.560&0.126&0.200--0.600\\
19&5&0.600 & 0.400&0.400&0.189&0.200--0.800\\
20&5&0.000 & 0.400&0.320&0.253&0.000--0.800\\
22&5&0.200 & 0.600&0.320&0.103&0.200--0.400\\
23&5&0.600 & 0.000&0.600&0.000&0.600--0.600\\
24&5&0.800 & 0.400&0.840&0.084&0.800--1.000\\
25&5&0.600 & 0.200&0.600&0.000&0.600--0.600\\
26&5&0.800 & 0.200&0.560&0.227&0.200--0.800\\
27&5&0.600 & 0.600&0.520&0.169&0.200--0.600\\
28&5&0.600 & 0.600&0.600&0.000&0.600--0.600\\
\midrule
Mean over datasets&--&0.5236&0.3764&\textbf{0.5402}&\multicolumn{2}{c}{$\mathrm{SD}_{\rm seed}=0.0187$}\\
\bottomrule
\end{tabular}}
\end{table*}

Complete seed-scheme values are reported in Appendix~\ref{app:seed-results}.

The dataset-level results differ widely.
Dataset~24 has the highest mean OQ ($0.840$), followed by datasets~8 and~9 ($0.780$).
These maxima are reported descriptively, for two reasons.
Eighteen of the 27 fMRI systems contain five scored reference pairs, so one correctly or incorrectly ordered pair changes OQ by $0.2$.
In addition, the largest values observed among 27 inspected datasets are weak evidence of general fMRI recovery.
At the other end, dataset~6 has a mean OQ of $0.318$ with relatively low seed-scheme dispersion.
Dataset~20 has a mean of the same order but the widest seed-scheme range in the benchmark,
\begin{equation}
    0\le\mathrm{OQ}_{20}^{(s)}\le0.8.
\end{equation}
Dataset~26 also ranges from $0.2$ to $0.8$, whereas the OQ values of datasets~7, 13, 23, 25 and~28 stay identical across the ten seed schemes.

The aggregate fMRI evidence therefore does not support a claim of uniform ordering recovery.
The evidence shows a near-chance pair-weighted score together with strong heterogeneity across datasets.
The macro-average moves little across seed schemes ($0.0187$), while the variation between datasets ($\approx0.141$) is much larger.
Reporting every seed-level value separates the benchmark-level average from the stability of each dataset-level OQ.

\subsection{Comparison with external lead-lag ordering}
\label{sec:external-leadlag}

The frozen full QS-TCN model is compared with the CCF-AUC ordering of \citet{bennett2022leadlag}.
The comparator is run with the authors' code, distance correlation, a maximum lag of five, and decreasing row means.
Each regime contains 100 systems with five variables, and both methods use identical 780-observation prefixes.
QS-TCN retains its ten-model ensemble, whereas the comparator requires no supervised training.
A volatility-oriented adaptation applied to squared standardized observations is evaluated as well.

\begin{table*}[t]
\centering
\caption{Mean ordering quality on synthetic systems. The linear lead-lag control is scored against factor-delay precedence.}
\label{tab:external-leadlag}
\small
\begin{tabular}{lccc}
\toprule
Regime & QS-TCN & Native dCor & Squared-input dCor \\
\midrule
Positive linear volatility & 0.8595 & 0.5740 & 0.7726 \\
Signed linear volatility & 0.7428 & 0.5475 & 0.7202 \\
Original SV mechanism & 0.5729 & 0.4643 & 0.5414 \\
Target-state nonlinear volatility & 0.7772 & 0.5633 & 0.6991 \\
Linear lead-lag control & 1.0000 & 1.0000 & 1.0000 \\
\bottomrule
\end{tabular}
\end{table*}

QS-TCN achieves a higher mean OQ than both tabulated comparators in all four volatility regimes.
The original SV mechanism is the hardest regime for all methods, with mean OQ between $0.4643$ and $0.5729$.
Table~\ref{tab:external-leadlag-diff} gives the paired differences, together with the number of systems in which QS-TCN scores higher, equal or lower than the comparator.

\begin{table*}[t]
\centering
\caption{Paired OQ differences (QS-TCN minus comparator) over the same 100 systems, with descriptive 95\% system-bootstrap intervals. W/T/L counts the systems in which QS-TCN scores higher, equal or lower.}
\label{tab:external-leadlag-diff}
\small
\begin{tabular}{llrrr}
\toprule
Regime & Comparator & Difference & 95\% interval & W/T/L \\
\midrule
Positive linear & Native dCor & $+0.2855$ & $[+0.2351,\,+0.3351]$ & 78/16/6 \\
& Squared-input dCor & $+0.0869$ & $[+0.0535,\,+0.1206]$ & 45/42/13 \\
\addlinespace
Signed linear & Native dCor & $+0.1953$ & $[+0.1441,\,+0.2462]$ & 67/14/19 \\
& Squared-input dCor & $+0.0226$ & $[-0.0291,\,+0.0758]$ & 35/33/32 \\
\addlinespace
Original SV & Native dCor & $+0.1086$ & $[+0.0457,\,+0.1714]$ & 53/16/31 \\
& Squared-input dCor & $+0.0314$ & $[-0.0229,\,+0.0857]$ & 44/23/33 \\
\addlinespace
Target-state nonlinear & Native dCor & $+0.2139$ & $[+0.1636,\,+0.2637]$ & 69/21/10 \\
& Squared-input dCor & $+0.0781$ & $[+0.0359,\,+0.1210]$ & 46/36/18 \\
\bottomrule
\end{tabular}
\end{table*}

Against the native ordering, all four intervals are positive.
Against squared-input dCor, the paired gains are $0.0869$ for positive linear coupling and $0.0781$ for target-state nonlinear coupling, and the other two intervals include zero.
The native Pearson, signature and lag-eight distance-correlation variants of the comparator also give positive intervals in all four volatility regimes.
On the control, QS-TCN and the distance-correlation and Pearson variants reach $1.0000$, and the signature variant reaches $0.9460$.
The estimates for these additional variants are provided in \path{results/summary.csv} and \path{results/paired_differences.csv} of the accompanying comparison archive.
The intervals are the 2.5th and 97.5th percentiles of the mean paired OQ differences obtained from 20,000 bootstrap resamples of systems.
The paired intervals are descriptive and are not adjusted for multiple comparisons.
The comparison remains exploratory, because the volatility benchmarks had been examined before.

\subsection{Forex and Gold markets}

\paragraph{Data.}
Daily closing prices are taken from the public GitHub repository \path{ejtraderLabs/historical-data}.
The repository provides one daily file for each of the six currency pairs and for gold quoted against the dollar (XAUUSD, labelled GOLD here).
Each file holds 2,400 daily records ending on 4 March 2022.
Only dates for which all seven series have a price are retained, which leaves 2,385 common dates from 4 December 2012 to 4 March 2022.
The returns are log-returns, $r_t=\log P_t-\log P_{t-1}$, computed between consecutive common dates, so a return over a weekend or a holiday covers more than one calendar day.
This gives $T=2384$ returns, from 5 December 2012 to 4 March 2022.
Zero returns occur in the panel, with 3 for EURUSD, 4 for GBPUSD, 5 for USDJPY, 3 for USDCHF, none for AUDUSD, 4 for USDCAD and 1 for GOLD.
These make 20 zeros among $16{,}688$ entries, each on a different date.
These dates are retained, because deleting every date with a zero would reduce the sample to $2364$ returns.

The QS-TCN pipeline is applied to an observed market panel with $T=2384$ observations of seven assets, namely EURUSD, GBPUSD, USDJPY, USDCHF, AUDUSD, USDCAD and GOLD.
Unlike Finance and fMRI, this experiment has no complete causal-order ground truth, so no reference-based accuracy score is reported.
The analysis reports the global order returned by the same inference mechanism as in the benchmarks and examines the strongest pairwise relations behind this order.

The full-series QS-TCN analysis returns
\begin{equation}
\mathrm{USDCHF}\prec\mathrm{USDJPY}\prec\mathrm{GOLD}\prec\mathrm{EURUSD}\prec\mathrm{GBPUSD}\prec\mathrm{AUDUSD}\prec\mathrm{USDCAD}.
\end{equation}
The pairwise field behind this order is uneven.
After bootstrap aggregation, the reported bootstrap weights differ across source-target relations.
Table~\ref{tab:forex-pairs} lists the strongest retained examples.

\begin{table}[t]
\centering
\caption{Selected directional relations from the frozen Forex/Gold QS-TCN analysis. $C_{ij}$ denotes the aggregated antisymmetric pairwise evidence and $R_{ij}$ its associated reported reliability.}
\label{tab:forex-pairs}
\scriptsize
\setlength{\tabcolsep}{5pt}
\renewcommand{\arraystretch}{0.94}
\begin{tabular}{lcc}
\toprule
Directional relation & $C_{ij}$ & $R_{ij}$\\
\midrule
GBPUSD $\rightarrow$ USDCAD & 0.0892 & 0.847\\
GOLD $\rightarrow$ GBPUSD & 0.0691 & 0.775\\
GOLD $\rightarrow$ USDCAD & 0.0958 & 0.759\\
USDCHF $\rightarrow$ USDCAD & 1.2944 & 0.621\\
USDCHF $\rightarrow$ EURUSD & 0.3230 & 0.589\\
USDJPY $\rightarrow$ GOLD & 0.1116 & 0.450\\
\bottomrule
\end{tabular}
\end{table}

This sequence is interpreted only as the order induced by the QS-TCN directional score and not as a validated causal graph.
The $R_{ij}$ values are the bootstrap weights reported with the archived pairwise scores, and they are not probabilities that a direction is correct.
The magnitude of $C_{ij}$ differs widely across relations, and this difference is not interpreted, because the sensitivity of the lag-asymmetry anchor to node scale is not assessed in the experiments reported in this section.
Without a complete market reference structure and without marginal-ordering controls, the Forex/Gold experiment is treated as an application of the frozen ordering mechanism.
The experiment should therefore not be read as validating either causal direction or independence from marginal scale.

\subsection{Cryptocurrency market comparisons}

Cryptocurrency markets provide a second real-data setting, in which the QS-TCN output is confronted with evidence produced independently of QS-TCN.
Ji et al.~\citep{ji2019} analyze return and volatility connectedness across six major cryptocurrencies.
Yousaf and Ali~\citep{yousafali2020} study return and volatility transmission among Bitcoin, Ethereum and Litecoin before and during the COVID-19 period.
These studies use different statistical objects, samples and modelling assumptions, so their reported directions are not treated as causal ground truth.
The comparison asks a narrower question, namely how much of the independently reported directional structure the QS-TCN order reproduces.
A published relation ``$a$ before $b$'' counts as reproduced when $a$ precedes $b$ in the QS-TCN order.
All permutation probabilities in this subsection are one-sided, and each gives the probability, under uniformly random orders of the assets, of at least the observed number of reproduced relations.

\begin{table*}[t]
\centering
\caption{Comparison of QS-TCN cryptocurrency orderings with directional structures reported in independent studies. Concordance is computed only for relations that map unambiguously onto the corresponding asset set.}
\label{tab:crypto-comparison}
\scriptsize
\setlength{\tabcolsep}{3pt}
\renewcommand{\arraystretch}{0.94}
\resizebox{0.96\textwidth}{!}{%
\begin{tabular}{p{2.3cm}p{4.1cm}p{6.2cm}p{3.5cm}}
\toprule
Study & Published comparison object & QS-TCN result & Agreement\\
\midrule
Ji et al.~\citep{ji2019} & 15 pairwise volatility-spillover directions and a separate net-spillover ranking & DASH $\prec$ BTC $\prec$ LTC $\prec$ XRP $\prec$ XLM $\prec$ ETH & Pairwise: $10/15\approx0.667$, $p_{\rm perm}=0.2347$; net ranking: $\tau=0.60$\\[2mm]
Yousaf--Ali~\citep{yousafali2020} & Regime-specific cryptocurrency transmission relations & Pre-COVID: ETH $\prec$ LTC $\prec$ BTC; COVID: BTC $\prec$ ETH $\prec$ LTC & $2/5$ pre-COVID; $2/3$ COVID; combined agreement $=4/8$\\
\bottomrule
\end{tabular}}
\end{table*}

\subsubsection{Comparison with Ji et al.}
Ji et al.~\citep{ji2019} examine connectedness through return and volatility spillovers across six large cryptocurrencies.
For the corresponding six-asset system, the frozen full-series QS-TCN ordering is
\begin{equation}
\mathrm{DASH}\prec\mathrm{BTC}\prec\mathrm{LTC}\prec\mathrm{XRP}\prec\mathrm{XLM}\prec\mathrm{ETH}.
\end{equation}
Among the 15 published pairwise directional relations used in the comparison, QS-TCN satisfies
\begin{equation}
    \frac{10}{15}\approx0.6667.
\end{equation}
The archived exact one-sided enumeration over the $6!=720$ possible orders, which counts the orders satisfying at least $10$ of the $15$ published relations, gives
\begin{equation}
    p_{\mathrm{perm}}=0.2347.
\end{equation}
The archived Kendall agreement with the separately reported net-spillover ranking is
\begin{equation}
    \tau=0.60.
\end{equation}
For two strict rankings of six assets, $\tau=(n_c-n_d)/15$, where $n_c$ and $n_d$ count the concordant and discordant pairs.
The value $\tau=0.60$ thus corresponds to $n_c=12$ and $n_d=3$.
The two agreement numbers describe different comparison objects and must not be mixed.
The value $10/15$ compares QS-TCN with the 15 published pairwise spillover directions, whereas $\tau=0.60$ compares the QS-TCN order with the separately reported net-spillover ranking.
The pairwise concordance is not unusual under exact random-order enumeration ($p_{\mathrm{perm}}=0.2347$).
The experiment is therefore interpreted as partial concordance between studies and not as independent validation of the QS-TCN ordering.

\subsubsection{Comparison with Yousaf and Ali}
Yousaf and Ali~\citep{yousafali2020} analyze high-frequency return and volatility transmission among Bitcoin, Ethereum and Litecoin, and they separate the pre-COVID-19 and COVID-19 periods.
This separation is retained and the complete sample is not pooled before inference, because the published study reports that transmission changes between the two regimes.

The directional constraints come from the cross-coefficients $b_{ij}$ of Table~3 of Yousaf and Ali.
A relation is retained only when its coefficient is significant at the $5\%$ level.
With the source-target convention stated in the table note, this rule gives five pre-COVID constraints, BTC$\to$ETH, BTC$\to$LTC, ETH$\to$BTC, ETH$\to$LTC and LTC$\to$ETH, and three COVID constraints, BTC$\to$ETH, BTC$\to$LTC and LTC$\to$ETH.
Negative coefficients keep their source-target orientation (Appendix~\ref{app:crypto-mapping}).
The input is the series of hourly log-returns.
Within each regime, the returns are split into consecutive non-overlapping windows of $1000$ returns, starting at the beginning of the regime, and the incomplete tail is dropped.
QS-TCN is run on every window under the ten frozen seeds.
The consensus ordering of a regime is the rank consensus over all window-seed pairs of that regime, so the assets are sorted by their mean position across these orders.
The two regimes are scored separately, and the final score is the sum of the two hit counts, $H=\mathrm{hits}_{\mathrm{pre\text{-}COVID}}+\mathrm{hits}_{\mathrm{COVID}}$.
The random-order law uses one independent uniform order per regime, which gives $6\times6=36$ equally likely joint outcomes.

For the pre-COVID period, ten complete 1,000-return windows are available, and the QS-TCN consensus ordering is
\begin{equation}
    \mathrm{ETH}\prec\mathrm{LTC}\prec\mathrm{BTC}.
\end{equation}
This order satisfies
\begin{equation}
    2/5
\end{equation}
of the pre-specified directional constraints retained from the published comparison object.
For the COVID period, two complete 1,000-return windows are available, and the recovered order changes to
\begin{equation}
    \mathrm{BTC}\prec\mathrm{ETH}\prec\mathrm{LTC},
\end{equation}
which satisfies
\begin{equation}
    2/3
\end{equation}
of the corresponding COVID-period constraints.
Combining the two regimes gives an overall agreement of
\begin{equation}
    \frac{4}{8}=0.50.
\end{equation}
The exact random-order distribution has the same median value, and the one-sided probability of at least four agreements under independent random orders in the two regimes is
\begin{equation}
    p_{\mathrm{perm}}=0.6667.
\end{equation}
The QS-TCN order changes between the pre-COVID and COVID windows.
This change is qualitatively compatible with regime-dependent transmission, but it does not validate the specific directional structure reported by Yousaf and Ali.
Across the eight retained directional constraints, the observed agreement is $4/8$, and the exact random-order comparison gives $p_{\mathrm{perm}}=0.6667$.
This result is therefore retained as a negative comparison and not as supporting evidence.

\clearpage
\appendix
\section{Supplementary Material: Seed-wise Benchmark Ordering Quality}
\label{app:seed-results}
The following tables report the ordering quality for each of the ten frozen bootstrap seed schemes (A--J).
They provide the per-scheme values underlying the benchmark summaries in Section~\ref{sec:results}.

\begin{table}[H]
\centering
\caption{Finance ordering quality for all ten frozen bootstrap seed schemes.}
\label{tab:finance-all-seeds}
\scriptsize
\setlength{\tabcolsep}{3.2pt}
\renewcommand{\arraystretch}{0.92}
\resizebox{0.94\textwidth}{!}{%
\begin{tabular}{lrrrrrrrrrrr}
\toprule
Dataset & A & B & C & D & E & F & G & H & I & J & Mean \\
\midrule
manyinputs &1.000&1.000&1.000&1.000&1.000&1.000&1.000&1.000&1.000&1.000&1.000\\
20\_1A &.778&.889&.833&.833&.778&.889&.722&.778&.778&.778&.806\\
20\_1B &.650&.550&.650&.600&.550&.650&.550&.500&.600&.600&.590\\
20\_1C &.526&.526&.526&.526&.579&.526&.526&.526&.526&.526&.532\\
20\_1D &.684&.737&.737&.684&.737&.684&.737&.632&.684&.737&.705\\
20\_1E &.824&.824&.647&.765&.824&.882&.647&.706&.882&.824&.782\\
40\_1 &.737&.711&.658&.763&.737&.737&.658&.737&.658&.605&.700\\
40\_1\_3 &.725&.650&.650&.700&.725&.700&.675&.775&.625&.775&.700\\
20\_1\_3 &.550&.600&.600&.600&.650&.650&.750&.650&.600&.600&.625\\
\bottomrule
\end{tabular}}
\end{table}

\begin{landscape}
\begin{table}[p]
\centering
\caption{fMRI ordering quality for every frozen bootstrap seed scheme. The table gives the ten recorded seed-scheme scores underlying the fMRI summaries in Table~\ref{tab:fmri-detailed}.}
\label{tab:fmri-all-seeds}
\scriptsize
\setlength{\tabcolsep}{2.4pt}
\renewcommand{\arraystretch}{0.80}
\resizebox{0.75\linewidth}{!}{%
\begin{tabular}{crrrrrrrrrrrr}
\toprule
DS&A&B&C&D&E&F&G&H&I&J&Mean&SD\\
\midrule
1&.400&.200&.600&.400&.200&.600&.600&.600&.600&.400&.460&.165\\
2&.545&.636&.455&.545&.636&.727&.818&.727&.636&.455&.618&.120\\
3&.444&.389&.444&.444&.444&.444&.444&.444&.389&.500&.439&.032\\
4&.475&.492&.525&.525&.475&.525&.492&.492&.525&.508&.503&.021\\
5&.400&.400&.400&.400&.400&.200&.400&.400&.400&.400&.380&.063\\
6&.273&.273&.273&.364&.273&.364&.364&.364&.364&.273&.318&.048\\
7&.400&.400&.400&.400&.400&.400&.400&.400&.400&.400&.400&.000\\
8&.800&.800&.800&.800&.800&.800&.800&.800&.800&.600&.780&.063\\
9&.800&.800&.800&.600&.800&.800&.800&.800&.800&.800&.780&.063\\
10&.600&.800&.600&.600&.800&.200&.600&.600&.600&.600&.600&.163\\
11&.545&.636&.545&.727&.727&.727&.455&.636&.727&.636&.636&.096\\
12&.727&.545&.455&.818&.818&.545&.636&.636&.727&.545&.645&.125\\
13&.625&.625&.625&.625&.625&.625&.625&.625&.625&.625&.625&.000\\
14&.400&.800&.600&.800&.600&.800&.600&.400&.800&.600&.640&.158\\
15&.600&.600&.600&.400&.600&.600&.600&.600&.400&.600&.560&.084\\
16&.429&.429&.429&.429&.429&.714&.429&.286&.143&.714&.443&.171\\
17&.273&.455&.545&.455&.455&.364&.273&.545&.455&.545&.436&.103\\
18&.600&.600&.600&.600&.200&.600&.600&.600&.600&.600&.560&.126\\
19&.600&.200&.400&.400&.400&.400&.200&.200&.400&.800&.400&.189\\
20&.000&.400&.200&.200&.000&.400&.800&.400&.600&.200&.320&.253\\
22&.200&.400&.400&.400&.400&.200&.400&.200&.200&.400&.320&.103\\
23&.600&.600&.600&.600&.600&.600&.600&.600&.600&.600&.600&.000\\
24&.800&.800&.800&.800&.800&1.000&.800&1.000&.800&.800&.840&.084\\
25&.600&.600&.600&.600&.600&.600&.600&.600&.600&.600&.600&.000\\
26&.800&.800&.400&.600&.200&.600&.600&.200&.800&.600&.560&.227\\
27&.600&.200&.200&.600&.600&.600&.600&.600&.600&.600&.520&.169\\
28&.600&.600&.600&.600&.600&.600&.600&.600&.600&.600&.600&.000\\
\bottomrule
\end{tabular}}
\end{table}
\end{landscape}
\clearpage

\section{Squared-memory modes}
\label{app:sv-modes}

Under the stationary Gaussian log-volatility model in Eq.~\eqref{eq:sv-reference}, let $v=\sigma_\eta^2/(1-\phi^2)$ and omit the node index.
Independent standard Gaussian measurement innovations, also independent of the log-volatility process, give $\mathbb E[X_t^2]=\exp(\mu+v/2)$.
Since $\operatorname{Cov}(h_t,h_{t-k})=v\phi^k$, the Gaussian exponential-moment identity yields, for $k\geq1$,
\begin{equation}
\begin{aligned}
\operatorname{Cov}(X_t^2,X_{t-k}^2)
&=\exp(2\mu+v)\bigl[\exp(v\phi^k)-1\bigr]\\
&=\exp(2\mu+v)\sum_{m=1}^{\infty}\frac{v^m}{m!}(\phi^m)^k.
\end{aligned}
\label{eq:squared-memory-expansion}
\end{equation}

Thus several geometric modes can arise from a single persistence parameter.
Channel~I characterizes these observable modes without assigning them a causal origin.
The restriction to positive lags is necessary, because the lag-zero expression also involves the fourth moment of the measurement innovation.

\section{Published Yousaf--Ali comparison set}
\label{app:crypto-mapping}

The comparison below measures agreement with published spillover estimates.
These estimates serve as comparison objects and not as causal ground truth.

The comparison retains all off-diagonal $b_{ij}$ coefficients in Table~3 of \citet{yousafali2020} with reported $p<0.05$.
Directions follow the table note, which interprets $b_{12}$ as BTC$\to$ETH, irrespective of coefficient sign.
The indexing in their variance equation suggests the opposite convention, so the reversal of all directions is also checked.
Printed $p$-values of $0.000$ are rounded values.

\begin{table}[htbp]
\centering
\caption{Retained Yousaf--Ali coefficients under the direction convention stated in their Table~3 note.}
\label{tab:yousaf-comparison-set}
\small
\renewcommand{\arraystretch}{1.12}
\begin{tabular}{llrr}
\toprule
Period & Direction & Coefficient & $p$-value \\
\midrule
Pre-COVID & BTC $\to$ ETH & $-0.144$ & 0.000 \\
Pre-COVID & BTC $\to$ LTC & $-0.129$ & 0.000 \\
Pre-COVID & ETH $\to$ BTC & $-0.058$ & 0.032 \\
Pre-COVID & ETH $\to$ LTC & $-0.269$ & 0.000 \\
Pre-COVID & LTC $\to$ ETH & $1.028$  & 0.000 \\
\addlinespace
COVID & BTC $\to$ ETH & $0.037$ & 0.000 \\
COVID & BTC $\to$ LTC & $0.025$ & 0.040 \\
COVID & LTC $\to$ ETH & $0.069$ & 0.000 \\
\bottomrule
\end{tabular}
\end{table}

The reported orders satisfy $2/5$ pre-COVID constraints and $2/3$ COVID constraints, giving $4/8$ overall.
Of the $6\times6=36$ pairs of independent uniform regime-specific orders, 24 achieve at least four agreements, which gives $p_{\mathrm{perm}}=0.6667$.
Reversing all directions gives $3/5$ and $1/3$, and the overall agreement is preserved.
Reciprocal relations limit the maximum combined agreement to $6/8$.

\section{Implementation details}
\label{app:implementation}

The settings below describe the reconstructed implementation used for the synthetic comparison.
They do not establish numerical equivalence with the historical benchmark implementation.

\subsection{Structured temporal filters}

Channel~I supplies the rates used by the structured filters in Channel~II.
The Memory Cell applies the normalized geometric filter
\begin{equation}
p_a(\ell)=
\frac{a^{\ell-1}}
{\sqrt{\sum_{s=1}^{L_m}a^{2(s-1)}}+10^{-8}},
\qquad \ell=1,\ldots,L_m.
\label{eq:geometric-memory-filter}
\end{equation}
This filter is applied at the target persistence $a=\widetilde\phi_j$ and at each accepted target pole $a=\widehat\lambda_{jr}$.
The readout combines these responses with free-filter responses and the descriptor $\Theta_j$.

For $i\to j$, the Heredity Cell filters the source history with Eq.~\eqref{eq:heredity-kernel}, which combines $\widetilde\phi_i$ and $\widehat\lambda_{jr}$.
Each kernel is normalized by its Euclidean norm, and its unnormalized first entry is one, including at coincident rates.
Unsupported pole slots are masked.
The estimated rates stay fixed, while the neural weights learn to combine the responses.

Both modules use eight-lag causal filters with unit dilation.
With one-based indexing, memory predictions use $t=9,\ldots,n$.
A second eight-sample truncation leaves $t=17,\ldots,n$ for directional pooling.

\subsection{Statistical pairwise score}
\label{app:statistical-score}

For each centered-square series $q_j$, autoregressions with an intercept are fitted.
Candidate orders range from zero to four and share the fitting window $t=5,\ldots,n$.
With $N=n-4$, the selected order minimizes
\begin{equation}
\mathrm{BIC}_{j,p}
=
N\log\!\left(\frac{\mathrm{RSS}_{j,p}}{N}\right)
+(p+2)\log N.
\end{equation}
The selected coefficients define the residuals
\begin{equation}
\zeta_{j,t}
=
q_{j,t}-\widehat c_j
-\sum_{\ell=1}^{\widehat p_j}\widehat a_{j,\ell}q_{j,t-\ell}.
\end{equation}
For $\mathcal I_k=\{\max(k,4)+1,\ldots,n\}$, the overlap-centered covariance is
\begin{equation}
\widehat\Gamma_{ij}(k)
=
\frac{1}{|\mathcal I_k|}
\sum_{t\in\mathcal I_k}
\bigl(q_{i,t-k}-\overline q_{i,k}\bigr)
\bigl(\zeta_{j,t}-\overline\zeta_{j,k}\bigr),
\end{equation}
where both means are computed on the same overlap.
The statistical score is
\begin{equation}
S_{0,ij}
=
\frac{1}{K_0}\sum_{k=1}^{K_0}
\left[\widehat\Gamma_{ij}(k)-\widehat\Gamma_{ji}(k)\right],
\qquad K_0=8.
\label{eq:statistical-score-explicit}
\end{equation}

These autoregressive residuals differ from the Memory Cell residuals in Eq.~\eqref{eq:memory}.
Autoregressions and covariances are refitted in every inference view.
Candidates with non-finite $\mathrm{RSS}$ or $\mathrm{RSS}\leq10^{-12}$ are excluded, and fitting stops if none remains.

\subsection{Bootstrap aggregation and decoding}
\label{app:bootstrap}

Inference uses $B=19$ synchronized circular-block views, each of length $n$, with block length $b_n=\max\{12,3\operatorname{round}(n^{1/3})\}$.
Uniformly sampled blocks are concatenated and truncated to $n$ indices, and the indices are shared across variables.
Centered squares are computed before resampling.
The Channel-I estimates and the neural parameters remain fixed, while $S_0$ is refitted.

The pairwise scores are aggregated as
\begin{align}
M_{ij}
&=\operatorname{median}_{b=1,\ldots,B}S_{ij}^{(b)}, \nonumber\\
V_{ij}
&=\operatorname{median}_{b=1,\ldots,B}
\left|S_{ij}^{(b)}-M_{ij}\right|, \nonumber\\
U_{ij}
&=\left|\frac{1}{B}\sum_{b=1}^{B}
\operatorname{sign}\!\left(S_{ij}^{(b)}\right)\right|, \nonumber\\
C_{ij}
&=U_{ij}
\exp\!\left[-\frac{V_{ij}}{2(|M_{ij}|+10^{-6})}\right]M_{ij},
\label{eq:bootstrap-explicit}
\end{align}
with $\operatorname{sign}(0)=0$.
The weight combines sign stability $U_{ij}$ and dispersion $V_{ij}$, and is not a probability of causal correctness.
Antisymmetry is enforced through $C\leftarrow(C-C^\top)/2$, with a zero diagonal.

For the synthetic comparison, the ten checkpoint-specific matrices are averaged before decoding.
Content hashes define a canonical column order for tie-breaking, and the labels are restored afterwards.
Starting from decreasing row sums, the decoder accepts single-node insertions with gain above $10^{-12}$, and stops when none improves the objective or after 25 moves.

\subsection{Training for the synthetic comparison}
\label{app:training-details}

The checkpoints used in Section~\ref{sec:external-leadlag} are trained on separate banks of 160 systems for memory and 160 for directional learning, with 80 independent validation systems.
Each bank covers the four volatility regimes equally.
Every system has five variables and 1,200 observations, and only the first 780 are used.
Each variable is centered and scaled by its empirical standard deviation on this prefix, with divisor $n$, before the centered squares are formed.
The Channel-I estimates are fixed during training.

The Memory Cell minimizes squared prediction error over valid times and variables and is frozen after six epochs.
For each synthetic graph, let $\mathcal P$ contain the ordered ancestor-descendant pairs.
The directional loss is
\begin{equation}
\mathcal L_{\mathrm{dir}}
=
\frac{1}{|\mathcal P|}
\sum_{(i,j)\in\mathcal P}
\log\!\left(1+\exp(-S_{ij}/0.25)\right)
+
\frac{0.02}{d^2}\|S-S_0\|_F^2.
\label{eq:training-explicit}
\end{equation}
Each optimizer step uses one system, and the systems are reshuffled at each epoch.
Training labels include ancestor pairs, whereas validation and test OQ use direct reference edges.

\begin{table}[htbp]
\centering
\caption{Architecture and optimization settings for the synthetic-comparison checkpoints. Memory parameters are frozen before directional training.}
\label{tab:synthetic-training-settings}
\small
\renewcommand{\arraystretch}{1.12}
\begin{tabular}{lll}
\toprule
Setting & Memory Cell & Directional module \\
\midrule
Lag horizon & 8 & 8 \\
Free filters & 12 & 12 \\
Parameters & 601 & 4,959 \\
Optimizer & AdamW & AdamW \\
Learning rate & $2\times10^{-3}$ & $1.5\times10^{-3}$ \\
Weight decay & $10^{-4}$ & $2\times10^{-4}$ \\
Epochs & 6 & 8 \\
Gradient-norm clipping & 5 & 5 \\
\bottomrule
\end{tabular}
\end{table}

The 5,560 parameters include 601 memory parameters, which are frozen during directional training.
Readout dimensions are $19\to24\to1$ for memory, $9\to24\to12$ for modal weights, $8\to16\to1$ for branch mixing, $56\to48\to24\to1$ for directional scoring, and $12\to16\to1$ for the symmetric gate.
Hidden layers use $\tanh$, and the mixing and gate outputs use sigmoid.
Free filters have independent Gaussian initialization with standard deviation $0.03$.
Linear layers use PyTorch initialization, and AdamW uses its default moment parameters.

Training seeds run from 97101 to 97110.
For each run, the checkpoint with the highest mean validation OQ is selected among epochs 1 to 8 after 19-view bootstrap aggregation and decoding, with the 80 systems weighted equally.
Exact ties retain the earliest epoch.
All ten checkpoints enter the ensemble.
Generator settings, data seeds and checkpoint hashes are recorded in the accompanying implementation.

\section{Conclusion}
\label{sec:conclusion}

This paper introduced QS-TCN, a framework that orders the variables of a multivariate time series from volatility memory. The method links statistical estimates of persistence and temporal modes to structured filters within a shared temporal network, then combines pairwise directional scores into a global order.
The historical benchmark evaluation yielded a mean OQ of 0.7155 on Finance and 0.5402 on fMRI. In a separate synthetic evaluation, the reconstructed implementation achieved higher mean OQ than the native lead--lag comparator in all four volatility regimes. Against the squared-input adaptation, descriptive bootstrap intervals for paired OQ differences excluded zero in two regimes. These comparisons assess the complete framework and do not isolate the contribution of individual components.
The recovered order describes precedence rather than direct causal links. Future work will examine the contribution of the structured filters and neural components, and evaluate QS-TCN as an ordering component for structural equation modelling.

\clearpage
\appendix


\end{document}